\pdfoutput=1

\documentclass[11pt]{article}

\usepackage[final]{acl}

\usepackage{times}
\usepackage{latexsym}

\usepackage[T1]{fontenc}

\usepackage[utf8]{inputenc}

\usepackage{microtype}

\usepackage{inconsolata}

\usepackage{amsmath}

\usepackage{graphicx}

\usepackage{url}

\usepackage{xcolor}

\definecolor{ourRed}{RGB}{255,150,146} 
\definecolor{ourBlue}{RGB}{38,159,255} 

\definecolor{customRed}{RGB}{209,41,32}
\definecolor{customBlue}{RGB}{46,89,167}

\usepackage{tabularx}

\usepackage{booktabs}
\usepackage{siunitx}
\usepackage{array}

\usepackage{multirow}

\title{Mind's Mirror:~Distilling Self-Evaluation Capability and\\Comprehensive Thinking from Large Language Models}

\author{
Weize Liu$^{1}$, Guocong Li$^{1}$, Kai Zhang$^{2}$, Bang Du$^{1}$, Qiyuan Chen$^{1}$, \\
\bf
Xuming Hu$^{3}$\thanks{\, Corresponding authors.}, \
Hongxia Xu$^{2,4}$\footnotemark[1], \
Jintai Chen$^{5}$, \
Jian Wu$^{2,4}$\\
$^{1}$Zhejiang University~~~
$^{2}$School of Public Health, Zhejiang University\\
$^{3}$The Hong Kong University of Science and Technology (Guangzhou)\\
$^{4}$Liangzhu Laboratory and Institute of Wenzhou, Zhejiang University\\
$^{5}$Computer Science Department, University of Illinois Urbana-Champaign\\
\texttt{\{weizeliu1115,xuminghu97\}@gmail.com} \ \ \ \texttt{einstein@zju.edu.cn}
}

\begin{document}
\maketitle
\begin{abstract}
Large language models (LLMs) have achieved remarkable advancements in natural language processing. However, the massive scale and computational demands of these models present formidable challenges when considering their practical deployment in resource-constrained environments. While techniques such as chain-of-thought (CoT) distillation have displayed promise in distilling LLMs into small language models (SLMs), there is a risk that distilled SLMs may still inherit flawed reasoning and hallucinations from LLMs. To address these issues, we propose a twofold methodology: First, we introduce a novel method for distilling the self-evaluation capability from LLMs into SLMs, aiming to mitigate the adverse effects of flawed reasoning and hallucinations inherited from LLMs. Second, we advocate for distilling more comprehensive thinking by incorporating multiple distinct CoTs and self-evaluation outputs, to ensure a more thorough and robust knowledge transfer into SLMs. Experiments on three NLP benchmarks demonstrate that our method significantly improves the performance of distilled SLMs, offering a new perspective for developing more effective and efficient SLMs in resource-constrained environments.
\end{abstract}

\section{Introduction}

\begin{figure*}[t]
  \centering
  \includegraphics[width=\textwidth]{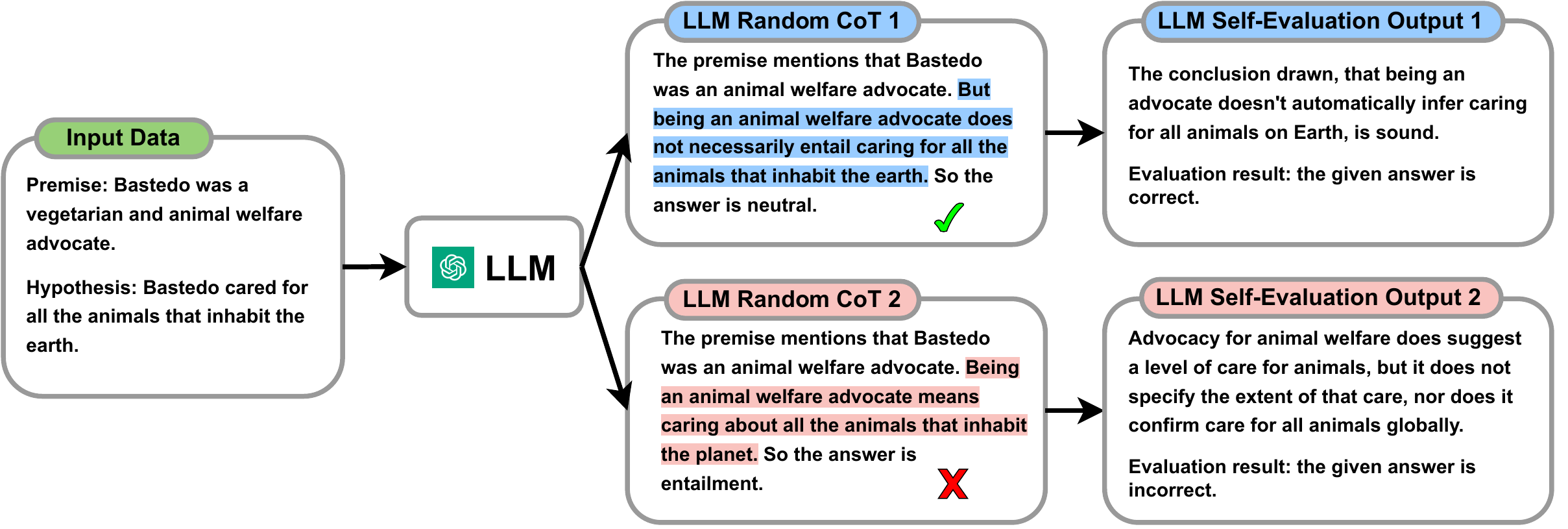}
  \caption{Examples of both the random CoT responses and their self-evaluation outputs generated by the LLM during natural language inference tasks. The human-like self-evaluation of the LLM enables the LLM to self-evaluate the correctness of its CoT reasoning processes, identifying which are correct (highlighted in \textcolor{ourBlue}{blue}) and which are incorrect (highlighted in \textcolor{ourRed}{red}) in these randomly generated CoT reasoning.}
  \label{fig:Picture_first}
  \vspace{-3mm}
\end{figure*}

With the gradual increase in the number of parameters, large language models (LLMs) have achieved significant successes in the field of natural language processing~\citep{brown2020language,kaplan2020scaling,hoffmann2022training,chowdhery2023palm,openai2023gpt4}. However, the tremendous model sizes and computational requirements of LLMs introduce challenges to their practical application, especially in resource-limited environments~\citep{zhao2023survey,zhu2023survey}. To address these challenges, various studies have delved into the compression of LLMs into small language models (SLMs) using knowledge distillation techniques and have led to significant reductions in computational complexity and inference costs~\citep{jiang-etal-2020-smart,gu2023knowledge,agarwal2023gkd}. This process involves traditional teacher-student learning methods and the more recent chain-of-thought (CoT) distillation method~\citep{zhu2023survey}. The CoT distillation methods use the CoT reasoning process of LLMs as supervision for training SLMs, rather than just labels. This allows SLMs to learn the reasoning process of LLMs, thereby improving the performance of SLMs.

While these CoT distillation methods have proven to be beneficial, they are not without their flaws, particularly:

\begin{enumerate}
    \item Even during the CoT distillation process, the distilled SLMs remain vulnerable to the flawed supervision provided by LLMs, as observations suggest that chains of thought (CoTs) generated by LLMs may contain hallucinations~\citep{zhang2023language}, accumulate errors~\citep{shen2021generate}, or lack robustness~\citep{vaswani2017attention,radford2019language,brown2020language,zhang2022opt}. As shown in the example in Figure~\ref{fig:Picture_first}, ``LLM Random CoT 2'' incorrectly broadens the scope of the premise by arguing that ``Being an animal welfare advocate means caring about all the animals that inhabit the planet.'' In practice, it is not easy to exclude these flawed CoTs, since the ground truth of CoTs is not always easily obtainable~\citep{zhang2023language}. Training SLMs with these flawed CoTs will result in SLMs inheriting these flaws and performance degradation~\citep{alemohammad2023self,ho-etal-2023-large}.
    \item A single instance of CoT might not capture the diverse reasoning routes LLMs can explore, limiting the richness of the distilled knowledge of SLMs. Furthermore, relying solely on the CoT reasoning process as supervision for training SLMs is insufficient to distill the comprehensive capabilities of LLMs, such as the ability to check the correctness of answers.
\end{enumerate}

To mitigate the impact of these flawed CoTs and allow SLMs to learn more comprehensive capabilities, we propose an innovative methodology that involves training SLMs to possess the self-evaluation capability. Humans often evaluate their reasoning processes to reduce errors in decision-making~\citep{poole2010artificial}, and a similar self-evaluation capability has also been observed in LLMs~\citep{kadavath2022language,shinn2023reflexion,madaan2024self,paul2023refiner}, which recognizes and corrects the generated hallucinations, faulty reasoning, and harmful content in a CoT~\citep{pan2023automatically}. Figure~\ref{fig:Picture_first} illustrates this with an example where incorrect reasoning in ``LLM Random CoT 2'' is identified and corrected in the self-evaluation. The advantage of self-evaluation is that it does not rely on external resources. However, it is constrained by the inherent capabilities of the model. To address this, we guide SLMs in distillation to learn the self-evaluation capability of LLMs. By learning the ability of LLMs to analyze right from wrong, SLMs can understand both what should and should not be generated, enhancing their predictive accuracy and reliability in various NLP tasks.

To facilitate comprehensive thinking and address the randomness and limitations of relying on a single CoT and a single self-evaluation, our second methodology insight involves distilling SLMs from diverse CoTs and multiple self-evaluation outputs generated by LLMs. This enables SLMs to inherit a broader range of comprehensive thinking capabilities since diverse CoTs and self-evaluation collectively offer a more comprehensive perspective, derived from the varied state spaces of LLMs.

In summary, our contributions can be outlined as follows:
\begin{enumerate}
    \item We distill the self-evaluation capability from LLMs into SLMs, which helps SLMs understand the potential reasons behind correct or incorrect reasoning and lays the foundation for mitigating errors (e.g., hallucinations) arising from flawed CoTs.
    \item We distill diverse CoTs and corresponding multiple self-evaluation outputs from LLMs into SLMs, enabling SLMs to learn a more comprehensive state space of LLMs. This approach empowers SLMs with enhanced reasoning and more comprehensive capabilities.
    \item Comprehensive experiments demonstrate that our method enables SLMs to inherit the self-evaluation capability and comprehensive thinking of LLMs, significantly enhancing the performance and reliability of distilled SLMs, and outperforming previous CoT distillation methods. This affirms our method is essential for creating robust and efficient SLMs capable of high-quality reasoning in resource-constrained environments.
\end{enumerate}

The code is available at \url{https://github.com/Attention-is-All-I-Need/Mind-s-Mirror-Distilling-LLM}.

\section{Related Work}

\paragraph{Chain-of-thought reasoning} Chain-of-thought (CoT) is a prompting method where a model generates intermediate reasoning steps to enhance its problem-solving capabilities~\citep{wei2022chain}. The chain-of-thought with self-consistency (CoT-SC)~\citep{Wang2023} builds upon CoT, sampling a set of diverse reasoning paths and selecting the most consistent answer as the final answer. This largely mitigates errors introduced by the inherent randomness of LLMs. The Tree of Thoughts (ToT) method~\citep{yao2024tree} models problem-solving as a tree search process, enabling LLMs to explore different reasoning pathways and conduct self-evaluation to determine the solution taken at each step. Therefore, by leveraging the capability of LLMs to generate diverse reasoning paths and self-evaluation, ToT significantly enhances the performance of LLMs in solving tasks such as Game of 24, Creative Writing, and Mini Crosswords.

\paragraph{Self-evaluation in LLMs} Many recent works have leveraged the self-evaluation capability of LLMs to enhance the reliability of their responses, such as Self-Refine~\citep{madaan2024self}, SelfCheck~\citep{miao2023selfcheck}, SelfCheckGPT~\citep{manakul-etal-2023-selfcheckgpt}, and Reflexion~\citep{shinn2023reflexion}. Concurrently, other studies have demonstrated the self-improvement potential of LLMs~\citep{huang-etal-2023-large,pan2023automatically}, as exemplified by RLAIF~\citep{lee2023rlaif}. However, these methods are designed for LLMs and do not consider distilling the self-evaluation capability into SLMs.

\paragraph{Knowledge distillation from LLMs}\label{para:knowledge_distillation} Knowledge distillation enhances the performance of smaller models by transferring knowledge from larger models~\citep{hinton2015distilling}. This method has been widely adopted for the optimization and compression of models. Recent studies have been focusing on leveraging the CoT reasoning generated by LLMs to enhance the performance of SLMs~\citep{wang-etal-2023-scott,magister-etal-2023-teaching,shridhar-etal-2023-distilling,wang-etal-2023-democratizing,chen-etal-2023-mcc,Fu2023SpecializingSL,zhu2023pad,saha2023can}. For instance, \citet{hsieh-etal-2023-distilling} introduced a ``Distilling step-by-step'' method for extracting rationales from LLMs as additional supervision for training SLMs. Similarly, \citet{li-etal-2023-symbolic} proposed the Symbolic Chain-of-Thought Distillation (SCoTD) method, which trains SLMs to learn CoT reasoning. Additionally, \citet{ho-etal-2023-large} presented ``Fine-tune-CoT'', a method that generates reasoning samples from LLMs to fine-tune SLMs. However, these methods do not consider mitigating the impact of harmful content in CoTs generated by LLMs on SLMs, as well as distilling other capabilities beyond CoTs. In contrast, our methodology incorporates the self-evaluation capability of LLMs into distillation, which can be utilized to mitigate the effects of flawed CoTs in a completely unsupervised manner and without relying on external resources, and allows SLMs to learn the more comprehensive capabilities of LLMs. Furthermore, some related works utilize SLMs with up to several billion parameters and have not been able to validate their effectiveness on SLMs with as few as 220M parameters, so our approach exhibits lower resource requirements and broader applicability.

\section{Distilling Self-Evaluation Capability and Comprehensive Thinking}

\begin{figure*}[th]
  \centering
  \includegraphics[width=\linewidth]{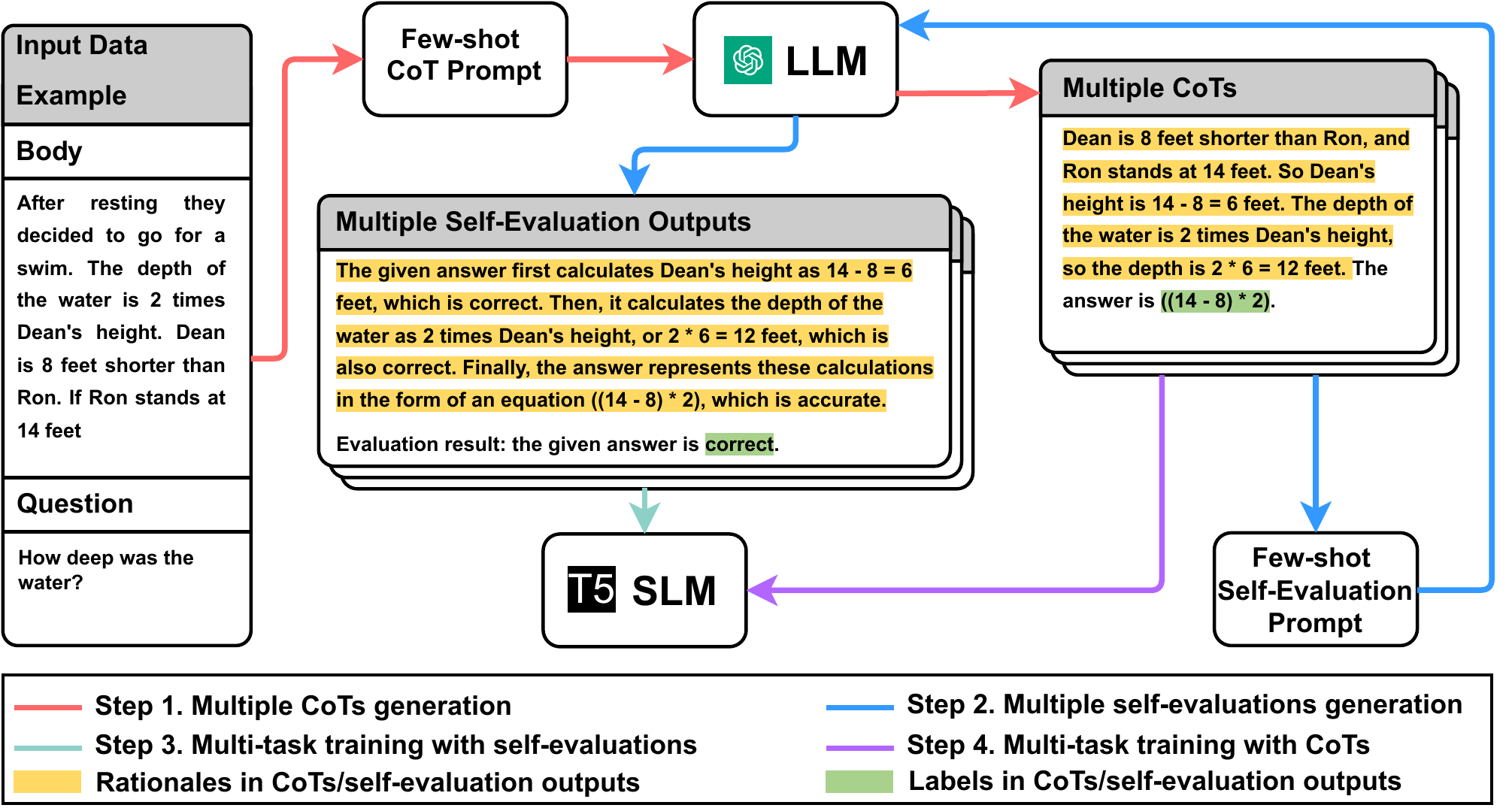}
  \caption{Detailed overview of our proposed methodology. \textbf{Step 1}: Obtain multiple CoTs from the LLM (Section~\ref{subsec:311}). \textbf{Step 2}: Obtain multiple self-evaluation outputs from the LLM (Section~\ref{subsec:312}). \textbf{Step 3}: Train the SLM with multiple self-evaluation outputs, enabling the SLM to distinguish right from wrong (Section~\ref{subsec:321}). \textbf{Step 4}: Train the SLM with multiple CoTs to give the SLM comprehensive reasoning capabilities (Section~\ref{subsec:322}).}
  \label{fig:picture_second}
  \vspace{-3mm}
\end{figure*}

We propose a new methodology for distilling the self-evaluation capability and comprehensive thinking of an LLM into an SLM. Our overall framework is illustrated in Figure~\ref{fig:picture_second}, which operates in 4 steps: (1) Given an LLM and an unlabeled dataset, we utilize CoT prompts to generate diverse rationales and corresponding pseudo-labels from the LLM. (2) By devising self-evaluation prompts, we enable the LLM to evaluate the correctness of its generated CoTs, which also include both the rationales and labels in its self-evaluation outputs. (3) Leveraging the rationales and labels in the self-evaluation outputs generated by the LLM, we employ multi-task learning to train the SLM, enabling the SLM to distinguish right from wrong. (4) Utilizing the diverse rationales in CoTs and labels from either LLM-generated pseudo-labels or human-annotated labels, we employ multi-task learning to train the SLM's reasoning capability.

\subsection{Obtaining diversity CoTs and self-evaluation outputs from the LLM}
\label{subsec:first}

In our pipeline, an LLM functions as the teacher, while an SLM serves as the student. First, we let the LLM generate multiple different CoTs and self-evaluation outputs for a given task. We utilize few-shot CoT prompting to enhance the quality and standardize the formats of the CoTs generated by the LLM. This process is shown as step 1 and step 2 in Figure~\ref{fig:picture_second}.

\subsubsection{Obtaining multiple CoTs}
\label{subsec:311}

For an unlabeled dataset \( D \), we devise a few-shot CoT prompt template \( p \) delineating how the task should be approached. We combine each input data \( x_i \) with \( p \) and use it as an input to LLM. With examples from \( p \), the LLM can simulate examples to generate the CoT response for \( x_i \) that contains a rationale \( r_i \) and a pseudo-label \( y_i \) (the yellow part and the green part of the ``Multiple CoTs Outputs'' in Figure~\ref{fig:picture_second}). We let the LLM regenerate several times to get multiple different CoTs.

\subsubsection{Obtaining multiple self-evaluation outputs}
\label{subsec:312}

After forming multiple CoTs representing different thoughts, a self-evaluation phase is initiated to evaluate the correctness of the CoTs. This is essential to imitate the complete human thought process and correct mistakes in reasoning. Given an unlabeled dataset \( D \), we devise a few-shot self-evaluation prompt template \( p_{\mathit{eval}} \), which guides the LLM in evaluating each CoT's correctness. For each CoT \( x_c \), shown in ``Multiple CoTs'' in Figure~\ref{fig:picture_second}, we add it to \( p_{\mathit{eval}} \) and use this as an input to prompt the LLM to generate the self-evaluation. With examples in \( p_{\mathit{eval}} \), the LLM simulates examples to generate the self-evaluation output for \( x_c \) that also contains a rationale \( r_{\mathit{eval_i}} \) and a label \( y_{\mathit{eval_i}} \) (the yellow part and the green part of the ``Multiple Self-Evaluation Outputs'' in Figure~\ref{fig:picture_second}).

Similarly, to distill a more comprehensive self-evaluation capability of the LLM, we generate multiple different self-evaluation outputs for each CoT. Multiple self-evaluation outputs along with multiple CoTs represent a more comprehensive and complete thought process for the LLM. Additionally, given the randomness of LLM outputs, we suggest examining the quality and diversity of multiple CoTs and self-evaluation outputs generated by the LLM for the same input, and removing duplicates and outputs of inferior quality, to enhance data quality. This is an optional step.

\subsection{Training the SLM with multiple self-evaluation outputs and diverse CoTs}
\label{subsec:second}

After generating diverse CoTs and their corresponding self-evaluation outputs using the LLM, we begin to train the SLM. Our training methodology for SLMs first emphasizes distilling self-evaluation capability to lay the foundation for reducing the impact of errors in CoTs on SLMs, followed by incorporating comprehensive reasoning capability through diverse CoTs distillation.~\citet{hsieh-etal-2023-distilling} have demonstrated that multi-task learning can lead to better performance than simply treating rationale and label predictions as a single joint task, and can reduce computation overhead during inference since it allows the SLM to directly predict labels without generating rationales. Hence, we employ multi-task learning to train the SLM for self-evaluation capability and CoT reasoning capability. By appending different ``task prefixes'' at the beginning of the input, we can direct the SLM to generate either a label or a rationale~\citep{raffel2020exploring}. We train the SLM to generate a label when the prefix is ``predict: '', and to generate a rationale when the prefix is ``explain: ''. This process is shown as step 3 and step 4 in Figure~\ref{fig:picture_second}.

\subsubsection{Distilling self-evaluation capability}
\label{subsec:321}

Using the self-evaluation data generated by the LLM, we aim to distill this capability into the SLM. During this phase, the model is trained to predict the self-evaluation label \( y_{\mathit{eval_i}} \) as well as generate corresponding rationale \( r_{\mathit{eval_i}} \). To guide the SLM in learning the self-evaluation outputs for each CoT, we employ a multi-task loss function:
\begin{align*}
L_{\mathit{SE}} &= \frac{1}{N_{\mathit{eval}}} \sum_{c=1}^{N_{\mathit{eval}}} \Big( \lambda \ell(f(x_c), y_{\mathit{eval}_c}) \\
&\quad + (1-\lambda) \ell(f(x_c), r_{\mathit{eval}_c}) \Big),
\end{align*}
where \( f \) represents the SLM and \( \ell \) is the cross-entropy loss between the tokens predicted by the SLM and the target tokens. \( x_c \) is the CoT that needs to be evaluated. \( \lambda \) is a hyperparameter for weighing the rationale loss. \( y_{\mathit{eval}_c} \) indicates the self-evaluation label generated by the LLM, \( r_{\mathit{eval}_c} \) is the rationale in the \(c^{th}\) self-evaluation output, and \( N_{\mathit{eval}} \) is the total amount of self-evaluation outputs.

\subsubsection{Distilling CoT reasoning capability}
\label{subsec:322}

After successfully distilling self-evaluation capability, the focus shifts to leveraging diverse CoTs to train the comprehensive reasoning capability of SLMs. For each instance in the dataset, we also employ a multi-task loss function to guide the SLM in learning CoT reasoning by:
\begin{align*}
L_{\mathit{CoT}} &= \frac{1}{N_{\mathit{CoT}}} \sum_{i=1}^{N_{\mathit{CoT}}} \Big( \lambda \ell(f(x_i), \hat{y}_i) \\
&\quad + (1-\lambda) \ell(f(x_i), r_{\mathit{CoT}_i}) \Big),
\end{align*}
where \( x_i \) indicates input data, \( \hat{y}_i \) indicates the pseudo-label \( y_i \) generated by the LLM or human-annotated label, \( r_{\mathit{CoT}_i} \) is the rationale in the \(i^{th}\) CoT, and \( N_{\mathit{CoT}} \) is the total amount of CoTs.

This two-pronged training regimen ensures that the SLM is not merely parroting the CoT reasoning but deeply understands introspective self-evaluation and nuanced reasoning, mirroring the powerful cognitive capabilities of the LLM.

\section{Experiments}

\paragraph{Tasks and datasets} To evaluate our distillation method, we conduct comprehensive experiments on three tasks: 1) math word problems (MWPs) task with the SVMAP dataset~\citep{patel-etal-2021-nlp}; 2) commonsense question answering (CQA) task with the CQA dataset~\citep{talmor2018commonsenseqa,rajani-etal-2019-explain}; 3) natural language inference (NLI) task with the ANLI dataset~\citep{nie-etal-2020-adversarial}. For dataset samples, we use either human-annotated labels from the dataset or LLM-generated pseudo-labels to explore the effect of human annotation availability on our method.

\paragraph{Setup} In distillation, we utilize gpt-3.5-turbo as the LLM\footnote{Most experiments were conducted in August 2023 using the gpt-3.5-turbo model provided by the OpenAI API.}. We utilize 5-shot CoT prompting to enhance the quality and standardize the formats of the responses generated by the LLM. We follow the CoT prompts from~\citet{wei2022chain} for the CQA dataset and devise similar prompts for other datasets and self-evaluation. To strike a balance between diversity and cost, in the main experiment, we obtain five CoTs for each training instance and five self-evaluation outputs of each CoT from the gpt-3.5-turbo model and choose the T5-Base model (220M)~\citep{raffel2020exploring} as the SLM. We provide more experimental details in Appendix~\ref{sec:appendix_a}. We also explore the effect of the value of the hyperparameter $\lambda$ on the results, which are presented in Appendix~\ref{sec:appendix_b}. Therefore, we select $\lambda = 0.5$ as the optimal hyperparameter for our main experiments. In all experiments, we report the mean results and standard deviations over 3 random runs.

\begin{table*}[ht!]
\centering
\small
\begin{tabular}{@{\hspace{5pt}}m{2.4cm}cccccc@{\hspace{5pt}}}
\toprule
\multirow{2}{*}{\textbf{Method}} & \multicolumn{2}{c}{\textbf{SVAMP}} & \multicolumn{2}{c}{\textbf{CQA}} & \multicolumn{2}{c}{\textbf{ANLI}} \\ 
\cmidrule(r){2-3} \cmidrule(r){4-5} \cmidrule(r){6-7}
& Pseudo-labels & Human-labels & Pseudo-labels & Human-labels & Pseudo-labels & Human-labels \\
\midrule
Standard \newline \scriptsize{Distillation / Fine-tuning} & 49.2 $\pm$ 1.9 & 59.3 $\pm$ 1.2 & 58.7 $\pm$ 0.4 & 62.0 $\pm$ 0.4 & 37.7 $\pm$ 1.2 & 42.1 $\pm$ 5.0 \\
\midrule
1 CoT \newline \scriptsize{(i.e., CoT distillation)}   & 51.7 $\pm$ 2.1 & 65.0 $\pm$ 1.1 & 59.7 $\pm$ 0.4 & 63.4 $\pm$ 0.2 & 39.8 $\pm$ 0.4 & 48.5 $\pm$ 1.2 \\
1 CoT \newline \scriptsize{w/ Self-Evaluation}   & 55.5 $\pm$ 0.4 & 67.8 $\pm$ 0.6 & 60.4 $\pm$ 0.2 & 63.7 $\pm$ 0.2 & 41.8 $\pm$ 0.4 & 49.2 $\pm$ 0.5 \\
\midrule
5 CoTs                            & 54.8 $\pm$ 1.0 & 68.7 $\pm$ 0.2 & 61.2 $\pm$ 0.4 & 63.9 $\pm$ 0.2 & 41.7 $\pm$ 0.4 & 49.7 $\pm$ 0.8 \\
5 CoTs \newline \scriptsize{w/ Self-Evaluation}  & \textbf{60.3 $\pm$ 0.6} & \textbf{72.7 $\pm$ 1.0} & \textbf{61.9 $\pm$ 0.3} & \textbf{65.0 $\pm$ 0.1} & \textbf{44.3 $\pm$ 0.2} & \textbf{50.8 $\pm$ 0.4} \\
\bottomrule
\end{tabular}
\caption{\textbf{Results of the main experiment.} We compare the accuracy (mean $\pm$ standard deviation, \%) of different distillation methods on three different datasets (SVAMP, CQA, and ANLI) using 220M T5-Base models, utilizing pseudo-labels generated by the LLM or human-annotated labels. The Human-labels represent human-annotated labels. The ``1 CoT'' adopts the ``Distilling step-by-step'' method proposed by~\citet{hsieh-etal-2023-distilling}.}
\label{tab:my-table}
\vspace{-3mm}
\end{table*}

\subsection{Main results}

Our results, presented in Table~\ref{tab:my-table}, show the advantages of our distillation method. Across all tasks and label types, the method we propose consistently outperformed the baselines (standard distillation and CoT distillation). In particular, we observe significant leaps in model performance when simultaneously training with five CoTs and their corresponding self-evaluation outputs. This reinforces our hypothesis about the value of incorporating self-evaluation and comprehensive thinking during the distillation process. Moreover, our approach exhibits a lower standard deviation than baseline methods, particularly under the ``5 CoTs w/ self-evaluation'' setting, indicating that our method offers stable improvements and enhances the robustness of distilled SLMs.

\paragraph{Effect of label quality} A discernible pattern from the results is the gap in performance between models trained using LLM-generated pseudo-labels and human-annotated labels. Given the typically higher accuracy of human-annotated labels, which are considered the gold standard in supervised learning, this result is expected. However, regardless of the type of training labels used, our method exhibits consistent advantages, suggesting that the benefits of our distillation method are also robust to variations in label quality.

\paragraph{Robustness across tasks} Our method's superiority is consistently evident when considering performance on different tasks, although the degree of improvement varies. In tasks such as MWPs (SVAMP dataset) and NLI (ANLI dataset), where reasoning complexity and potential for hallucinatory content are higher, the benefits of our methodology are more pronounced. This suggests that the proposed method effectively mitigates flawed reasoning and hallucinations in complex reasoning scenarios. In tasks like CQA (CQA dataset), where the reasoning processes might be less convoluted, the increments in performance are smaller yet still notable. This showcases the adaptability of our method to different types of reasoning complexity within various NLP tasks.

\subsection{Effect of model size}
\label{subsec:model_size}

To analyze the effectiveness of our proposed method across different model sizes, we further conducted experiments on the SVAMP dataset using both the T5-Small (60M) and T5-Large (770M) models. The results are presented in Table~\ref{tab:model_size-table}. Our method shows significant performance improvements on models of different sizes, reflecting the robustness of our method to model scale.

\begin{table*}[ht!]
\centering
\small
\begin{tabular}{@{\hspace{5pt}}m{2.4cm}cccccc@{\hspace{5pt}}}
\toprule
\multirow{2}{*}{\textbf{Method}} & \multicolumn{2}{c}{\textbf{T5-Small}} & \multicolumn{2}{c}{\textbf{T5-Base}} & \multicolumn{2}{c}{\textbf{T5-Large}} \\ 
\cmidrule(r){2-3} \cmidrule(r){4-5} \cmidrule(r){6-7}
& Pseudo-labels & Human-labels & Pseudo-labels & Human-labels & Pseudo-labels & Human-labels \\
\midrule
Standard \newline \scriptsize{Distillation / Fine-tuning} & 25.5 $\pm$ 1.7 & 30.8 $\pm$ 1.6 & 49.2 $\pm$ 1.9 & 59.3 $\pm$ 1.2 & 60.2 $\pm$ 1.5 & 76.5 $\pm$ 1.2 \\
\midrule
1 CoT \newline \scriptsize{(i.e., CoT distillation)}   & 29.2 $\pm$ 1.4 & 32.5 $\pm$ 0.4 & 51.7 $\pm$ 2.1 & 65.0 $\pm$ 1.1 & 66.2 $\pm$ 1.2 & 77.0 $\pm$ 1.2 \\
1 CoT \newline \scriptsize{w/ Self-Evaluation}   & 37.2 $\pm$ 1.4 & 35.2 $\pm$ 1.2 & 55.5 $\pm$ 0.4 & 67.8 $\pm$ 0.6 & 68.0 $\pm$ 1.1 & 79.0 $\pm$ 0.4 \\
\midrule
5 CoTs                            & 36.5 $\pm$ 2.0 & 33.3 $\pm$ 1.0 & 54.8 $\pm$ 1.0 & 68.7 $\pm$ 0.2 & 66.5 $\pm$ 0.7 & 81.3 $\pm$ 0.8 \\
5 CoTs \newline \scriptsize{w/ Self-Evaluation}  & \textbf{39.3 $\pm$ 1.2} & \textbf{36.8 $\pm$ 0.8} & \textbf{60.3 $\pm$ 0.6} & \textbf{72.7 $\pm$ 1.0} & \textbf{69.3 $\pm$ 0.6} & \textbf{83.7 $\pm$ 0.6} \\
Performance Gain                            & \textcolor{red}{+ 10.1} & \textcolor{red}{+ 4.3} & \textcolor{red}{+ 8.6} & \textcolor{red}{+ 7.7} & \textcolor{red}{+ 3.1} & \textcolor{red}{+ 6.7} \\
\bottomrule
\end{tabular}
\caption{\textbf{Experimental results for models of different sizes.} ``Performance Gain'' refers to the improvement in performance of our proposed method (``5 CoTs w/ Self-Evaluation'') relative to the baseline method (``1 CoT'').}
\label{tab:model_size-table}
\vspace{-3mm}
\end{table*}

\subsection{Effect of the number of CoTs}
\label{subsec:numCoT}

\begin{figure}[t]
  \centering
  \includegraphics[width=\linewidth]{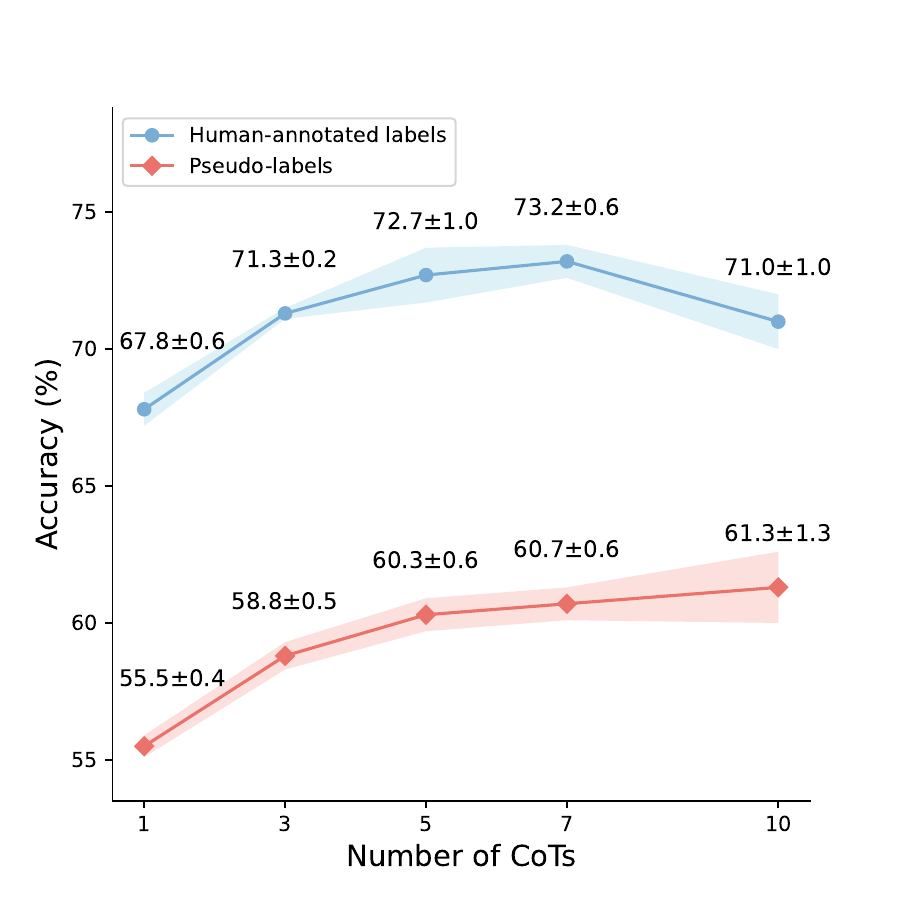}
  \caption{The experimental results of our method using the T5-Base model on the SVAMP dataset for different numbers of CoTs.}
  \label{fig:CoT_number}
  \vspace{-4mm}
\end{figure}

Using the SVAMP dataset as an example, we explore the effect of varying the number of CoTs on our method, where each CoT is accompanied by five self-evaluation outputs. As shown in Figure~\ref{fig:CoT_number}, initially, as the number of CoTs increases from 1 to 5, there is a notable improvement in performance metrics across both pseudo-labels and human-annotated labels. This trend underlines the benefit of exposing SLMs to a broader spectrum of reasoning processes and self-evaluation outputs, enhancing their capability to navigate complex reasoning and correct flawed reasoning. SVAMP as math word problems may benefit from a variety of different solutions, CQA as commonsense question answering may acquire richer knowledge from different answers, and ANLI as natural language inference might also benefit from different explanations. However, diminishing returns are observed when the number of CoTs exceeds five. In particular, when the number of CoTs exceeded 7, performance degradation is observed using human-annotated labels. It indicates that while multiple CoTs and self-evaluation outputs enrich the model's reasoning capabilities, there is a threshold beyond which performance cannot be further enhanced. This could be attributed to several factors: one possibility is that the integration of too many CoTs may introduce noise or conflicting reasoning patterns, thereby disrupting the distilled SLM. Another factor could be the cognitive load on the SLM. Beyond a certain scope, the model may struggle to effectively learn from additional training data.

This observation underscores the importance of finding an optimal balance in the number of CoTs used for distillation. As the number of CoTs and self-evaluation outputs increases, there is a corresponding rise in data costs and training expenses. Therefore, we opted to use five CoTs in our main experiments, balancing cost and performance.

\begin{figure}[ht]
  \centering
  \includegraphics[width=\linewidth]{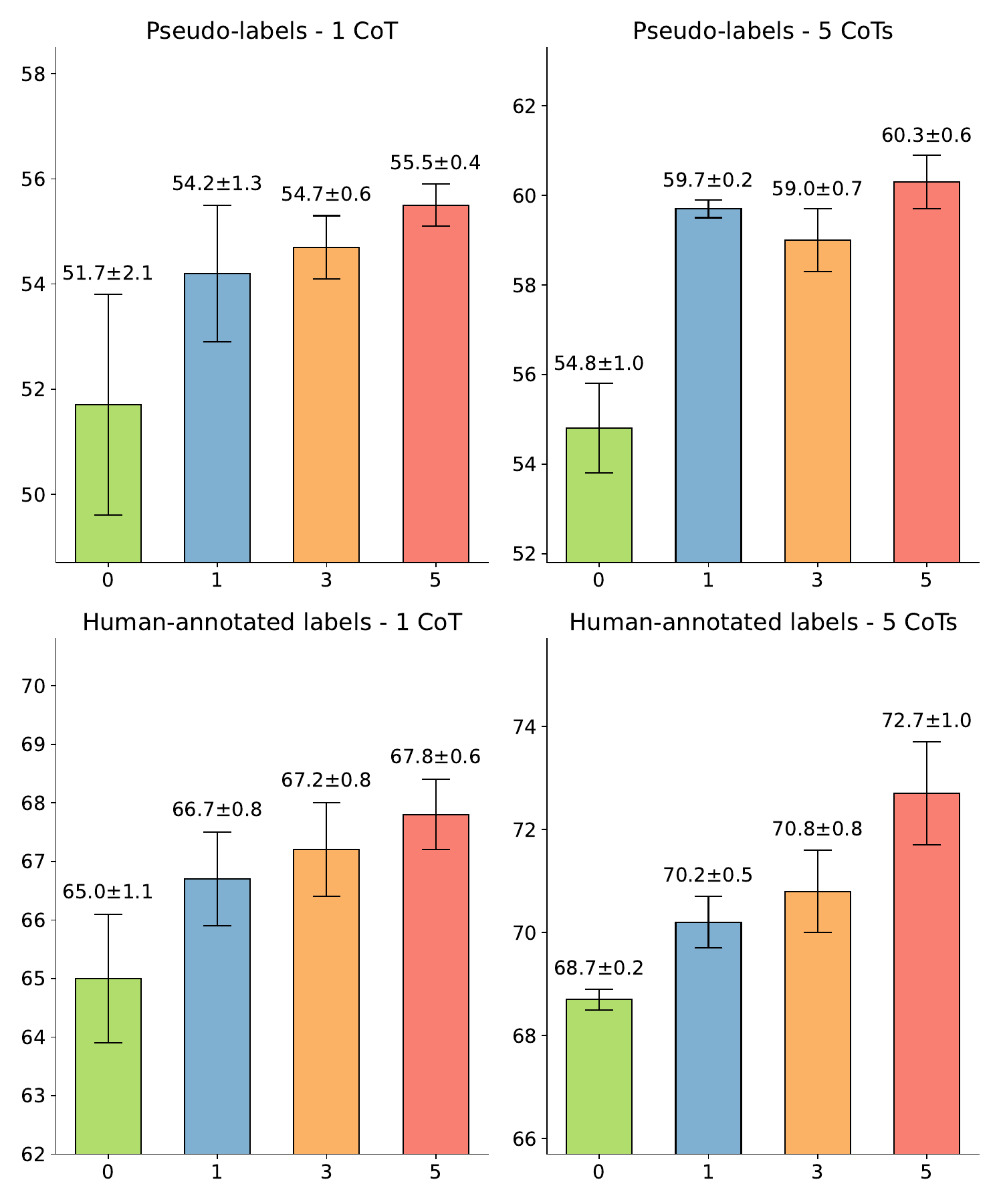}
  \caption{We present experimental results on the SVAMP dataset using the T5-Base model for different numbers of self-evaluation outputs for each CoT. Experiments are conducted under the settings of a single CoT and five CoTs, respectively.}
  \label{fig:self-evaluation_number}
  \vspace{-4mm}
\end{figure}

\begin{table*}[!ht]
\centering
\small
\begin{tabular}{@{\hspace{5pt}}m{3cm}|m{12cm}@{\hspace{5pt}}}
\toprule
\multicolumn{2}{c}{ANLI Example\hspace{0.8cm}Ground Truth Label: neutral} \\ \midrule
Model Input & Premise: East Bengal Football Club is a professional football club, based in Kolkata, West Bengal, India. It currently competes in the I-League, the top tier of Indian football. The club has won three National Football League (India) (later named as I league) titles, eight Federation Cups, and three Indian Super Cups, among others. The club is the current holder of the Calcutta Football League. \\
& Hypothesis: All of the team members live in West Bengal. \\ \midrule
Model Output \newline
(CoT Distillation) & Rationale: The premise mentions that East Bengal Football Club is based in Kolkata, West Bengal, India. \textcolor{customRed}{The hypothesis states that all of the team members live in West Bengal. The hypothesis is consistent with the information provided in the premise.} \\
& Label: entailment \\ \midrule
Model Output \newline
(Our Method) & Rationale: The premise mentions that East Bengal Football Club is based in Kolkata, West Bengal, India. \textcolor{customBlue}{However, it does not provide any information about the location of the team members. Therefore, the premise neither supports nor contradicts the hypothesis.} \\
& Label: neutral \\ \bottomrule
\end{tabular}
\caption{\textbf{A case study on the ANLI dataset.} The model trained by the CoT distillation method incorrectly predicts the label as entailment due to the premise location matching the hypothesis statement (highlighted in \textcolor{customRed}{red}), while the model trained by our method correctly identifies the lack of information regarding the team members' residences and correctly predicts the label as neutral (highlighted in \textcolor{customBlue}{blue}).}
\label{tab:case1}
\vspace{-3mm}
\end{table*}

\subsection{Effect of the number of self-evaluation outputs for each CoT}
\label{subsec:numSE}

We also investigate how the comprehensiveness of the self-evaluation affects the performance of our method by varying the number of self-evaluation outputs for each CoT. As shown in Figure~\ref{fig:self-evaluation_number}, in both pseudo-labels and human-annotated labels settings, we can observe that: as the number of self-evaluation outputs per CoT increases, there is a notable enhancement in the accuracy of the distilled SLMs, although it may not strictly be monotonically increasing. This indicates that distilling more self-evaluation outputs enables SLMs to produce more accurate and reliable outputs. Notably, accuracy improves more with five CoTs than with a single CoT, underscoring the synergistic effect of combining multiple CoTs with corresponding self-evaluation. Overall, these findings emphasize the importance of incorporating self-evaluation in the distillation and confirm the value of introspective self-evaluation in improving the reasoning and predictive capabilities of SLMs. Such introspective capabilities enable models to refine internal representations, rectifying possible misconceptions or potential pitfalls in their reasoning.

\section{Discussion}

\subsection{Can our method mitigate the flawed reasoning and hallucinations of SLMs?}

We conduct case studies on three datasets in the setting of using pseudo labels generated by LLMs. In the ANLI dataset case presented in Table~\ref{tab:case1}, the task is to judge the relationship between the premise and hypothesis. The model trained by the CoT distillation method incorrectly infers that the premise entails the hypothesis because superficially the geographic locations mentioned in the two statements match each other. This flawed reasoning likely results from a lack of critical evaluation of the information's depth and relevance, a pitfall in models trained without a self-evaluation mechanism. Conversely, the model trained by our method identifies the lack of specific information about team members' residences in the premise and correctly concludes that the premise is neutral to the hypothesis. This accurate judgment showcases our method's strength in instilling a comprehensive and critical reasoning capability in the model, enabling it to discern the nuances and gaps in information that affect the reasoning. Case studies on other datasets are in Appendix~\ref{sec:appendix_c}. The results indicate that our method effectively reduces flawed reasoning and hallucinations produced by distilled SLMs.

Given the absence of a gold standard for quantifying model hallucinations or harmful content, each of our 10 researchers (all holding Bachelor's degrees or higher) examined the outputs of different models for 200 pieces of data (with corresponding compensation). They manually compared the occurrences of hallucinations or harmful content in the outputs of models trained using our method and models trained using the CoT distillation baseline method. We statistically found that, on average, in approximately 7\% of the cases, models trained with our method exhibited a significant reduction in hallucinations or harmful content, 91\% of the cases tied and less than 2\% contained more hallucinations or harmful content.

\subsection{Can distilled SLMs really learn the self-evaluation capability?}

Previous works (refer to Section~\ref{para:knowledge_distillation}) have already demonstrated that SLMs can achieve CoT reasoning by learning from the CoTs generated by teacher models. Based on this, we propose that SLMs should also be able to master a certain level of self-evaluation capability through learning from the self-evaluation outputs generated by teacher models. \citet{gudibande2023false} point out that ``distilled imitation models are adept at mimicking ChatGPT's style but not its factuality'', because crowd workers rate their outputs as competitive with ChatGPT, yet their performance on NLP benchmarks does not improve. However, our paper demonstrates through tests on three NLP benchmarks that our method significantly improves the performance of distilled SLMs. Therefore, the SLMs distilled by our method do not merely imitate the style of ChatGPT, but indeed enhance the model's capabilities. Furthermore, our study improves the capability of imitation models by using extensive imitation data in situations of limited resources and unchangeable base SLMs, which is consistent with the approach given by \citet{gudibande2023false} to improve the capability of imitation models.

In Appendix~\ref{sec:appendix_d}, we tested SLMs trained with self-evaluation capability for their accuracy in evaluation predictions and printed their evaluation outputs. The results indicate that SLMs trained with self-evaluation capability achieve a consistency rate of approximately 90\% with GPT-3.5 evaluations and are capable of producing rational evaluation processes. In contrast, SLMs without self-evaluation training were completely unable to perform evaluations.

\begin{table}[ht!]
\centering
\small
\begin{tabular}{@{\hspace{5pt}}m{2.5cm}cc@{\hspace{5pt}}}
\toprule
\textbf{Method} & \multicolumn{2}{c}{\textbf{Reduced CQA}} \\
\cmidrule(r){2-3}
& Pseudo-labels & Human-labels \\
\midrule
Standard \newline \scriptsize{Distillation / Fine-tuning} & 41.6 $\pm$ 3.4 & 46.7 $\pm$ 1.2 \\
\midrule
1 CoT \newline \scriptsize{(i.e., CoT distillation)}   & 45.1 $\pm$ 1.2 & 47.1 $\pm$ 1.5 \\
1 CoT \newline \scriptsize{w/ Self-Evaluation}   & 42.6 $\pm$ 2.0 & 45.9 $\pm$ 1.3 \\
\midrule
5 CoTs                            & 44.8 $\pm$ 0.6 & 48.9 $\pm$ 1.6 \\
5 CoTs \newline \scriptsize{w/ Self-Evaluation}  & \textbf{46.1 $\pm$ 0.1} & \textbf{49.0 $\pm$ 0.6} \\
\bottomrule
\end{tabular}
\caption{The experimental results of training using 900 samples from the CQA dataset.}
\label{tab:reduced-cqa}
\vspace{-3mm}
\end{table}

\subsection{What leads to differences in effectiveness?}

Compared to CQA and ANLI, our method shows greater effectiveness on smaller SVAMP dataset. Is this due to the diminishing returns of our method as the volume of training data increases? We select the CQA dataset, which shows the least performance gain in our experiments, and reduce the number of training samples used from 8,766 to 900 to match the scale of SVAMP (keeping the test set unchanged) and then conduct experiments. The experimental results are presented in Table~\ref{tab:reduced-cqa}. Under the full training sample setting of the CQA dataset, ``5 CoTs w/ Self-Evaluation'' provides a performance gain of 2.2\% and 1.6\% respectively compared to ``1 CoT'' under two labels. In the setting of 900 training samples, the performance gains are 1.0\% and 1.9\% respectively. For the SVAMP dataset, the performance gains are 8.6\% and 7.7\%. Therefore, we believe that the returns of our method do not diminish with the increase in training data volume, but are more closely related to the nature of different tasks. SVAMP, as a math word problems task, is more likely to benefit from feedback through self-evaluation, while CQA, as a commonsense question answering task, benefits less. However, in our experiments, regardless of the task type, our method proved effective, demonstrating the universality of our approach.

\begin{table}[ht!]
\centering
\small
\begin{tabular}{@{\hspace{5pt}}m{2.5cm}cc@{\hspace{5pt}}}
\toprule
\textbf{Method} & \multicolumn{2}{c}{\textbf{SVAMP}} \\
\cmidrule(r){2-3}
& Pseudo-labels & Human-labels \\
\midrule
5 CoTs                            & 54.8 $\pm$ 1.0 & 68.7 $\pm$ 0.2 \\
5 CoTs \newline \scriptsize{w/ Self-Evaluation}  & 60.3 $\pm$ 0.6 & \textbf{72.7 $\pm$ 1.0} \\
\midrule
10 CoTs                            & 55.8 $\pm$ 1.0 & 67.6 $\pm$ 0.2 \\
10 CoTs \newline \scriptsize{w/ Self-Evaluation}  & \textbf{61.3 $\pm$ 1.3} & 71.0 $\pm$ 1.0 \\
\bottomrule
\end{tabular}
\caption{The experimental results of expanding the number of distilled CoTs to 10 CoTs on the SVAMP dataset.}
\label{tab:cot-10}
\vspace{-3mm}
\end{table}

\subsection{Can learning self-evaluation be replaced by learning more CoTs?}

From Table~\ref{tab:cot-10}, it can be observed that the marginal gain of increasing from ``5 CoTs'' to ``10 CoTs'' is almost negligible, and the performance of ``10 CoTs'' is significantly lower than that of ``5 CoTs w/ Self-Evaluation''. In the case of ``10 CoT'', the incorporation of self-evaluation distillation still manages to enhance the performance of the model. Therefore, we further confirmed that the role of self-evaluation cannot be substituted by merely adding more CoT data. When increasing the number of CoTs is ineffective, employing our proposed method of distilling with self-evaluation can further enhance model performance, breaking through the performance ceiling of CoT distillation.

\section{Conclusion}

In this study, we have introduced an innovative method to effectively distill the more comprehensive capabilities from LLMs into SLMs, emphasizing both the transfer of self-evaluation capability and comprehensive thinking, to mitigate the shortcomings of previous CoT distillation methods. Comprehensive experiments demonstrate that our method outperforms prior distillation methods consistently in various NLP tasks, significantly improving the performance and reliability of SLMs. We hope that this study can promote the more effective and efficient utilization of SLMs, especially in resource-limited environments.

\section{Limitations}

Despite the promising results and advancements achieved in our study, certain limitations need acknowledgment and further investigation:

\begin{enumerate}
\item \textbf{Limited teacher and student models}: The experiments we conducted primarily utilized a single teacher model, GPT-3.5, and two student models, T5-Base and T5-Large. While these selections were influenced by their current popularity and efficacy, it is crucial to note that the landscape of LLMs and SLMs is rapidly evolving. As such, our distillation method may manifest differently when paired with other architectures or models. Future work will involve testing a wider range of models to confirm the universality of our method.

\item \textbf{Limited tasks}: Although we evaluated our methods on three different NLP tasks, NLP tasks are broad and complex. Therefore, future evaluations of our method's performance on a wider range of tasks are needed to provide a more comprehensive evaluation of its strengths and potential weaknesses.

\item \textbf{Self-evaluation reliability}: One inherent limitation of the self-evaluation process is its reliance on the LLM's capacity for introspection. If the LLM's self-evaluation mechanism is flawed or biased, it might adversely affect the distilled SLM. In future work, we will investigate the differences in self-evaluation capabilities among different LLMs, such as Llama 2~\citep{touvron2023llama}, GPT-3.5, and GPT-4~\citep{openai2023gpt4}, and how these differences affect the performance of distilled SLMs.
\end{enumerate}

In conclusion, while we have made significant strides in advancing the distillation process from LLMs to SLMs, there exists a plethora of avenues for further refinement and exploration. Future endeavors should aim to address these limitations to ensure broader and more robust applicability.

\section{Ethical Considerations}

\paragraph{Potential risks} While our approach is dedicated to reducing the flaws inherited by SLMs from LLMs, SLMs may still inherit harmful biases and discrimination from LLMs. Therefore, future work will aim to further minimize the impact of harmful content from LLMs on SLMs.

\paragraph{The use of closed source LLMs} Many related studies and open source models have already utilized data obtained from the GPT family of models provided by OpenAI. We also obtain CoTs and self-evaluation outputs from the gpt-3.5-turbo model. However, the purpose of this study is not to develop models that compete with general large language models like ChatGPT. Instead, it aims to enhance the effectiveness and efficiency of small language models in resource-constrained environments, promoting the democratization of NLP. We only use gpt-3.5-turbo as the LLM to validate the effectiveness of our method, and it is not required to use the gpt-3.5-turbo model in practical applications, so different LLMs can be employed according to the licenses.

\paragraph{The use of AI assistants} We employed ChatGPT to assist us in polishing our paper and writing code.

\section*{Acknowledgements}

This research was partially supported by the Key R\&D Program of Zhejiang under grant No.~2024SSYS0026.

\bibliography{anthology,custom}

\clearpage
\appendix

\section{Experimental details}
\label{sec:appendix_a}

\paragraph{Datasets} The dataset statistics are shown in Table~\ref{tab:dataset}. Following~\citet{hsieh-etal-2023-distilling}, for the SVAMP dataset, 20\% of the original data is used as the test set. For the CQA dataset, the original validation set is used as the test set. Then, for both datasets, 10\% of the data from the original training set is sampled to serve as the validation set. The ANLI dataset follows the original split. The language of all datasets is English. To the best of our knowledge, all datasets used have been widely employed in NLP research and do not contain any information that names or uniquely identifies individual people or offensive content.

\begin{table}[h]
\centering
\begin{tabular}{lccc}
\toprule
Dataset & Train & Validation & Test \\
\midrule
SVAMP & 720 & 80 & 200 \\
CQA & 8,766 & 975 & 1,221 \\
ANLI & 16,946 & 1,000 & 1,000 \\
\bottomrule
\end{tabular}
\caption{Dataset statistics.}
\label{tab:dataset}
\end{table}

\paragraph{LLM performance} In Table~\ref{tab:llm}, we report the accuracy of LLM (gpt-3.5-turbo) on three datasets in our experiments, including accuracy on the training set (i.e., the accuracy of pseudo-labels used for training SLMs) and accuracy on the test set.

\begin{table}[h]
\centering
\begin{tabular}{lccc}
\toprule
Dataset & SVAMP & CQA & ANLI \\
\midrule
Training Set & 85.6 & 69.1 & 68.6 \\
Test Set & 84.3 & 72.4 & 55.1 \\
\bottomrule
\end{tabular}
\caption{The accuracy (\%) of LLM (gpt-3.5-turbo).}
\label{tab:llm}
\end{table}

\paragraph{Models \& Training} The T5-Small\footnote{\url{https://huggingface.co/google/t5-v1_1-small}} (60M), T5-Base\footnote{\url{https://huggingface.co/google/t5-v1_1-base}} (220M) and T5-Large\footnote{\url{https://huggingface.co/google/t5-v1_1-large}} (770M) models are all initialized with pre-trained weights obtained from Hugging Face, and the hyperparameter settings for their training are shown in Table~\ref{tab:model}. We perform the main experiments on 4 A100 GPUs.

\begin{table}[ht]
\centering
\begin{tabular}{lccc}
\toprule
Hyperparameter & \small{T5-Small / T5-Base} & T5-Large \\
\midrule
Total Batch Size & 64 & 32 \\
Learning Rate & $5 \times 10^{-5}$ & $5 \times 10^{-5}$ \\
Max Input Length & 1,024 & 1,024 \\
Maximum Steps & \multirow{2}{*}{4,000} & \multirow{2}{*}{9,000} \\
\small{(for SVAMP)} & & \\
Maximum Steps & \multirow{2}{*}{12,000} & \multirow{2}{*}{-} \\
\small{(for CQA \& ANLI)} & & \\
\bottomrule
\end{tabular}
\caption{Training hyperparameter settings.}
\label{tab:model}
\end{table}

\begin{figure}[!ht]
  \centering
  \includegraphics[width=\linewidth]{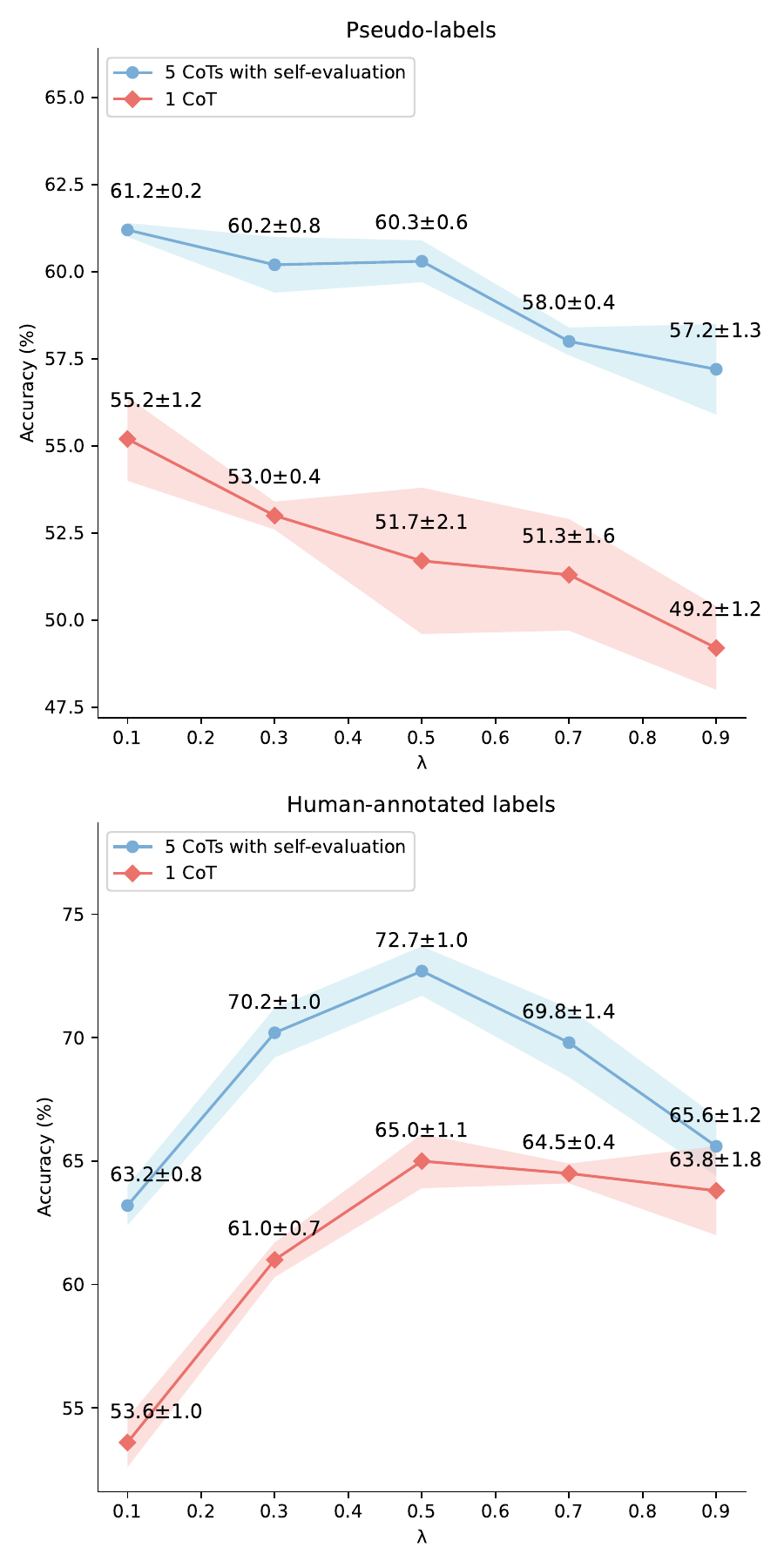}
  \caption{We present experimental results of distillation using the T5-Base model on the SVAMP dataset with different $\lambda$ values for ``1 CoT'' and ``5 CoTs with self-evaluation'' respectively.}
  \label{fig:lambda}
  \vspace{-3mm}
\end{figure}

\section{Effect of the hyperparameter \texorpdfstring{$\lambda$}{lambda}}
\label{sec:appendix_b}

As shown in Figure~\ref{fig:lambda}, our experiments reveal trends regarding the effect of the hyperparameter $\lambda$ on the accuracy of the SLMs trained using both pseudo-labels and human-annotated labels. 

For pseudo-labels, the performance of both methods declines as $\lambda$ increases, yet our approach exhibits a lesser decrease. Contrastingly, in the case of human-annotated labels, we observe a different trend. The accuracy initially increases with $\lambda$, peaking at $\lambda = 0.5$, and then begins to decline. This pattern underscores a critical observation: up to a certain point ($\lambda \leq 0.5$), increasing the weight on human-annotated labels positively impacts the model's ability to predict labels. However, beyond this optimal point, overly emphasizing human-annotated labels while neglecting rationales can lead to a decrease in label prediction accuracy. This suggests that the best way to enhance model performance is to learn high-quality labels and rationales in a balanced way. The differing trends observed between pseudo-labels and human-annotated labels may be attributed to variations in label quality: human-annotated labels, being of higher quality, benefit the model's accuracy when their weight is increased, whereas low-quality pseudo labels do not require higher weighting.

Based on these observations, we select $\lambda = 0.5$ as the optimal hyperparameter for our main experiments, maintaining a balance between the weights of labels and rationales.

\section{Case study}
\label{sec:appendix_c}

The detailed case studies presented in Tables~\ref{tab:case1},~\ref{tab:svamp_case}, and~\ref{tab:cqa_case} provide insightful examples demonstrating the effectiveness of our methodology compared to the baseline CoT distillation method. These cases highlight the importance of incorporating both self-evaluation and comprehensive thinking in the distillation process, which significantly reduces flawed reasoning and hallucinations in SLMs.

In the SVAMP example (Table~\ref{tab:svamp_case}), the model trained by the baseline CoT distillation method exhibits flawed reasoning in its calculation, erroneously summing the hours for learning Chinese and Spanish only, resulting in an incorrect total. This illustrates a common issue with CoT distillation, where the model may focus on a part of the problem, leading to incomplete reasoning. In stark contrast, the model trained by our method correctly identifies and sums the hours for all three languages, demonstrating a more comprehensive understanding and accurate reasoning process. This accurate reasoning underscores the effectiveness of our method, which incorporates both multiple CoTs and self-evaluation capability. By exposing the model to diverse reasoning processes and enabling it to evaluate its reasoning, our method equips the model to consider all relevant information comprehensively and to avoid flawed reasoning paths.

Similarly, in the CQA example (Table~\ref{tab:cqa_case}), the model trained by the baseline CoT distillation method incorrectly concludes that the most logical result of dying is a change of color, showcasing a case of flawed reasoning and hallucination. This error is likely due to a superficial association between the concepts of dying and color change, without a deeper understanding of the context of organic material decay. The model trained by our method, on the other hand, correctly identifies ``death and decay'' as the logical result of dying in the context of organic material, reflecting a deeper and more accurate comprehension of the question's essence. The comprehensive thinking instilled by our method, coupled with the ability to critically evaluate its reasoning, enables the model to select the most logical answer from the provided choices.

\section{Self-evaluation prediction accuracy and evaluation outputs of SLMs}
\label{sec:appendix_d}

Table~\ref{tab:eval_acc} shows that the evaluation results of SLMs (T5-Base) are mostly consistent with those of GPT-3.5, indicating that they have learned most of the evaluation capabilities of GPT-3.5. Table~\ref{tab:eval_case} illustrates an example where the SLM trained with self-evaluation capability can conduct reasonable evaluation, whereas that without this training is unable to evaluate CoT. The outputs of SLMs in other samples are similar to the examples shown.

\begin{table}[h]
\centering
\begin{tabular}{ccc}
\toprule
SVAMP & CQA & ANLI \\
\midrule
84.7 & 94.1 & 91.8 \\
\bottomrule
\end{tabular}
\caption{The self-evaluation prediction accuracy (\%) of SLMs (T5-Base) trained with self-evaluation capability on three datasets.}
\label{tab:eval_acc}
\end{table}

\begin{table*}[!ht]
\centering
\small
\begin{tabular}{@{\hspace{5pt}}m{3cm}|m{12cm}@{\hspace{5pt}}}
\toprule
\multicolumn{2}{c}{SVAMP Example\hspace{0.8cm}Ground Truth Label: ( ( 6.0 + 7.0 ) + 3.0 )} \\ \midrule
Model Input & Every day Ryan spends 6 hours on learning English 7 hours on learning Chinese and 3 hours on learning Spanish. How many hours does he spend on learning English, Chinese, and Spanish in all? \\ \midrule
Model Output \newline
(CoT Distillation) & Rationale: Ryan spends 6 hours on learning English and 7 hours on learning Chinese. \textcolor{customRed}{To find out how many hours he spends on learning English, we need to add the number of hours spent on learning Chinese and Spanish together: 7 + 3 = 6.} \\
& Label: (7 + 3) \\ \midrule
Model Output \newline
(Our Method) & Rationale: Ryan spends 6 hours on learning English, 7 hours on learning Chinese, and 3 hours on learning Spanish in total. \textcolor{customBlue}{To find the total number of hours he spends on learning English, Chinese, and Spanish in all, we need to add the number of hours spent on learning English, Chinese, and Spanish together. So, the equation is 6 + 7 + 3.} \\
& Label: (6 + 7) + 3 \\ \bottomrule
\end{tabular}
\caption{\textbf{A case study on the SVAMP dataset.} The model trained by the CoT distillation method incorrectly calculates the total time to learn the three languages (highlighted in \textcolor{customRed}{red}), while the model trained by our method correctly sums the time to learn the three languages (highlighted in \textcolor{customBlue}{blue}).}
\label{tab:svamp_case}
\end{table*}

\begin{table*}[!ht]
\centering
\small
\begin{tabular}{@{\hspace{5pt}}m{3cm}|m{12cm}@{\hspace{5pt}}}
\toprule
\multicolumn{2}{c}{CQA Example\hspace{0.8cm}Ground Truth Label: death and decay} \\ \midrule
Model Input & Question: As a result of dying, what happens to organic material? \\
& Choices: [change of color, stop breathing, wake up, death and decay, getting cold] \\ \midrule
Model Output \newline
(CoT Distillation) & Rationale: The question is asking about the result of dying. Of the given choices, \textcolor{customRed}{the most logical result of dying is a change of color.} \\
& Label: change of color \\ \midrule
Model Output \newline
(Our Method) & Rationale: The question is asking about the result of dying and the process of organic material. Of the given choices, \textcolor{customBlue}{the most logical result of dying is death and decay, as organic material undergoes the process of decay.} \\
& Label: death and decay \\ \bottomrule
\end{tabular}
\caption{\textbf{A case study on the CQA dataset.} The model trained by the CoT distillation method incorrectly considers that the most logical result of dying is a change of color (highlighted in \textcolor{customRed}{red}), while the model trained by our method correctly identifies the most logical result as death and decay (highlighted in \textcolor{customBlue}{blue}).}
\label{tab:cqa_case}
\end{table*}

\begin{table*}[!ht]
\centering
\small
\begin{tabular}{@{\hspace{5pt}}m{3cm}|m{12cm}@{\hspace{5pt}}}
\toprule
\multicolumn{2}{c}{SVAMP Self-Evaluation Example\hspace{0.8cm}Ground Truth Label: correct} \\ \midrule
Model Input & Paige and her mom planted 10 seeds in each flower bed, and there are 45 flower beds. So, they planted a total of 10 * 45 = 450 seeds. The answer is (10 * 45). \\ \midrule
Model Output \newline
(w/o Self-Evaluation) & Rationale: . The answer is (10 * 45). So, they planted 450 seeds. So, they are 450 seeds. So, they are 450 seeds. So, they are 450 seeds. So, they are 450 seeds. So, they are. So. So. So. So. So. So \\
& Label: - \\ \midrule
Model Output \newline
(w/ Self-Evaluation) & Rationale: Evaluation: The given answer correctly states that Paige and her mom planted 10 seeds in each flower bed and there are 45 flower beds. Then, it calculates the total number of seeds planted as 10 * 45 = 450 seeds, which is accurate. Finally, the answer represents these calculations in the form of an equation (10 * 45), which is correct. \\
& Label: correct \\ \bottomrule
\end{tabular}
\caption{\textbf{A case study on the evaluation of a CoT for the SVAMP dataset by SLMs trained with self-evaluation capability versus those not trained with self-evaluation capability.} The SLM trained with self-evaluation capability can output rationales and labels that are coherent and well-reasoned. In contrast, SLMs that are not trained with self-evaluation capability fail to produce logically sound rationales and formally appropriate labels.}
\label{tab:eval_case}
\end{table*}

\end{document}